\pdfoutput=1

\documentclass[11pt]{article}

\usepackage[]{acl}

\usepackage{times}
\usepackage{latexsym}

\usepackage[T1]{fontenc}

\usepackage[utf8]{inputenc}

\usepackage{microtype}

\usepackage{inconsolata}

\usepackage{bbm}
\usepackage{threeparttable}
\usepackage{booktabs, multirow, xspace}
\usepackage{todonotes}
\newcommand{\Note}[2]{} 
\newcommand{\SideNote}[2]{}
\renewcommand{\Note}[2]{\todo[color=#1,size=\small, inline=true]{#2}} 
\renewcommand{\SideNote}[2]{\todo[color=#1,size=\small]{#2}} %

\newenvironment{itemizesquish}{\begin{list}{\labelitemi}{\setlength{\itemsep}{-0.2em}\setlength{\labelwidth}{0.5em}\setlength{\leftmargin}{\labelwidth}\addtolength{\leftmargin}{\labelsep}}}{\end{list}}

\usepackage{amsmath}
\DeclareMathOperator*{\argmax}{arg\,max}

\newcommand{\ambigls}{ambiguous label set}
\newcommand{\ambiglst}{\mathcal{L}_{ambig, t}}

\newcommand{\randguess}{\textsc{freq}\xspace}
\newcommand{\base}{\textsc{zero}\xspace}
\newcommand{\staticn}{\textsc{static-$N$}\xspace}
\newcommand{\topretr}{\textsc{retr}\xspace}
\newcommand{\std}[1]{{\tiny $\pm$#1}}

\title{Ambiguity-Aware In-Context Learning with Large Language Models}

\author{Lingyu Gao\thanks{~~Work done as an intern at Google Research.} \\
{\small Toyota Technological Institute at Chicago} \\
{\small \texttt{lygao@ttic.edu}}
\And
Aditi Chaudhary \\
{\small Google Research} \\
{\small \texttt{aditichaud@google.com}}
\And
Krishna Srinivasan \\
{\small Google Research} \\
{\small \texttt{krishnaps@google.com}}
\AND
Kazuma Hashimoto \\
{\small Google Research} \\
{\small \texttt{kazumah@google.com}}
\\\And
Karthik Raman  \\
{\small Google Research} \\
{\small \texttt{karthikraman@google.com}}
\\\And
Michael Bendersky \\
{\small Google Research} \\
{\small \texttt{bemike@google.com}}
}

\begin{document}
\maketitle
\begin{abstract}

In-context learning (ICL), i.e., showing large language models (LLMs) only a few task-specific demonstrations, has led to downstream gains without task-specific fine-tuning.
However, LLMs are sensitive to the choice of prompts, and therefore a crucial research question is how to select good demonstrations for ICL.
One effective strategy is leveraging semantic similarity between the ICL demonstrations and test inputs by using a text retriever, which however is sub-optimal as that does not consider the LLM's existing knowledge about that task.
From prior work \cite{lyu-etal-2023-z}, we already know that labels paired with the demonstrations bias the model predictions.
This leads us to our hypothesis whether \emph{considering LLM's existing knowledge about the task, especially with respect to the output label space can help in a better demonstration selection strategy}.
Through extensive experimentation on three text classification tasks, we find that it is beneficial to not only choose semantically similar ICL demonstrations but also to choose those demonstrations that help resolve the inherent label ambiguity surrounding the test example.
Interestingly, we find that including demonstrations that the LLM previously mis-classified and also fall on the test example's decision boundary, brings the most performance gain. 
\end{abstract}

\section{Introduction}

Leveraging LLMs \citep{DBLP:conf/nips/BrownMRSKDNSSAA20, DBLP:journals/corr/abs-2204-02311, DBLP:journals/corr/abs-2201-08239}  via \emph{in-context learning} (ICL) is now a popular strategy for improving downstream task performance, wherein the model is able to perform a task by simply being conditioned on the task definition and/or few task \emph{demonstrations} (input-output examples) \cite{DBLP:conf/nips/BrownMRSKDNSSAA20,xie2021explanation}.

\begin{figure*}[t!]
\begin{center}
\includegraphics[width=\textwidth]{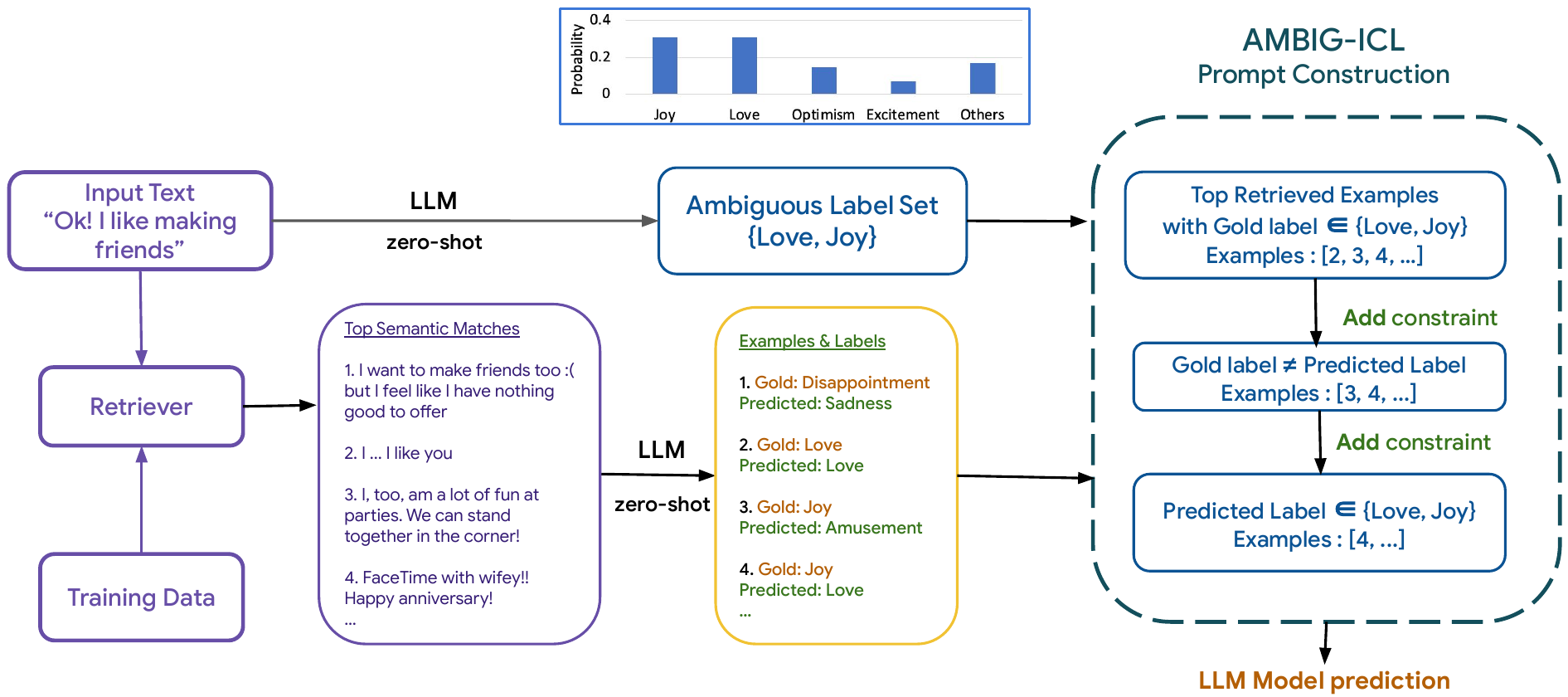}
\end{center}
\caption{\label{fig:sample-figure}Overview of our proposed method for selecting ICL demonstrations: For each test example, we first use a retriever to rank training data by semantic similarity. At the same time, we identify the ambiguous label set for each test example and also obtain the output predictions on the retrieved training data. 
Next, we apply three constraints on the top-ranked demonstrations which are: 1) select those demonstrations whose gold label is in the ambiguous label set, 2) select those which are also mis-classified by the model, and 3) select those mis-classified examples whose predicted label is in the ambiguous label set. 
Finally, we construct prompts with selected ICL demonstrations to get the final model predictions. 
} 
\end{figure*}

As ICL gets increasingly adopted, it has brought to light \citep{lester-etal-2021-power, liu-etal-2022-makes, zhang-etal-2022-active, lu-etal-2022-fantastically}
that LLMs are sensitive to the choice of prompts, making ``prompt engineering'' for different tasks challenging and time-consuming.
However, prompt engineering does not have to be a complete guessing game; rather it can be governed by some data-derived signals.
For example, selecting demonstrations that are semantically similar to a new input has shown to be more effective over randomly sampled demonstrations~\cite{das-etal-2021-case,liu-etal-2022-makes,margatina2023active}, wherein a text retriever is used to select the top-$k$ training examples for each test example based on the \emph{input text}.
The motivation is that using information from existing similar situations will help solve a new problem \citep{DBLP:journals/aicom/AamodtP94}.

\begin{figure}[!t]
\begin{center}
\includegraphics[width=0.95\linewidth]{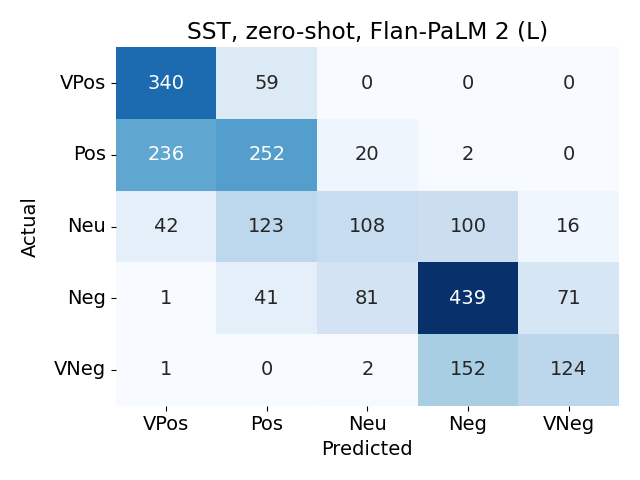}
\end{center}
\caption{\label{fig:cm_L}Confusion Matrix of zero-shot experiments on SST with Flan-PaLM 2 (L). Labels: VPos (Very Positive), Pos (Positive), Neu (Neutral), Neg (Negative), VNeg (Very Negative).
} 
\end{figure}

However, the solely input-based selection does not explicitly capture the LLM's existing knowledge about the task-specific \emph{label space} of both the ICL demonstration as well as the test input.
For example, on a five-way sentiment classification task (SST~\citep{socher-etal-2013-recursive}), we have observed that the Flan-PaLM 2 model (size L) \cite{DBLP:journals/corr/abs-2305-10403} is confused between two specific labels, `Very Negative' and `Negative,' a lot more than say between `Neutral' and `Very Negative', as shown in \autoref{fig:cm_L}.
This motivates us to investigate whether \emph{the model's existing knowledge can also be leveraged to select even more effective demonstrations.}

Specifically, we derive signals from the underlying LLM about the output label space of both the new test example and the training data from which we select the demonstrations.
As motivated above, the model's ambiguity around the new test example's output label will help us know \emph{what the model is most confused about}, which in turn can be used to select those demonstrations that help reduce this confusion.
For selecting such demonstrations from the training data, we propose to consider not only the ground truth labels paired with these demonstrations, but also the \emph{usefulness} by looking at their model prediction. 
First, given a test example and pool of training data, for each test example we use an off-the-shelf retriever to retrieve top-$k$ examples that have similar input text.
For each test example, we identify an \emph{ambiguous label set} of two output labels that the model is most confused about.
Next, we select top-ranked demonstrations such that their ground truth labels lie in the above label set.
To further find \emph{useful} demonstrations, we identify those which are mis-classified by the model;
the intuition is that showing the model a previously mis-classified demonstration could force it to correct it \citep{DBLP:journals/eswa/Tan06, DBLP:conf/aaai/WangZWFSM20}.
Finally, on top of the mis-classified demonstrations we add a constraint to select only those demonstrations whose model prediction falls within the ambiguous label set, i.e., on the test example's decision boundary.

To test our hypothesis, we focus on multi-class text classification tasks that have fine-grained nuance in the label space.
We conduct extensive experimentation across three   tasks, namely SST \cite{socher-etal-2013-recursive}, GoEmotions \cite{demszky-etal-2020-goemotions}, and EDOS (Task-B) \cite{kirk-etal-2023-semeval}, all of which have fine-grained label space,  making the model more likely to be confused across labels. Our key observations are:

\begin{enumerate}
    \item Incrementally adding constraints, i.e., 1) considering \emph{label ambiguity of test example}, 2) limiting ICL demonstrations to \textit{mis-classified demonstrations}, and 3) considering \emph{label ambiguity of training examples} leads to +1.5\%, +2.2\%, +2.6\% improvement in F1 macro scores over the retriever-based ICL, averaged across all datasets (\autoref{tab:f1_diff}).
    \item We find that adding such label-based constraints helps more on a smaller model, i.e., on Flan-PaLM 2 (M) (+3.9\% gain) compared to +1.4\% gain on Flan-PaLM 2 (L).
    \item We also attribute this success of our proposed methods to the observation that the \emph{ambiguous label set acts as a good proxy to the gold test label}, and as noted by \citet{min-etal-2022-rethinking}, labels in the ICL demonstrations bias the model predictions the most. 
    Therefore, showing the models the `likely' gold label guides the model to make the correct prediction (\autoref{gold_label_num}).

\end{enumerate}

\section{Proposed Method} \label{sec:method}
Typically, in an ICL regime, we assume access to training data $\mathcal{D}_{train}=\{(x_0, y_0), \cdots, (x_T, y_T)\}$ from which the goal is to select $d$ demonstrations to be used as prompt.
As motivated in the introduction, we follow a three-step  approach for selecting  demonstrations, for each test example, we need to 1) extract semantically similar examples from $\mathcal{D}_{train}$, 2) identify the ambiguous label-set and 3) extract model predictions for $\mathcal{D}_{train}$ to identify mis-classified examples.
Below, we describe each step in more detail and how they are used together to select the ``best'' demonstrations.

\paragraph{Extract Semantically Similar Demonstrations}
Typically, in this approach, demonstrations are selected for each test example $x_t$ by finding those examples from the $\mathcal{D}_{train}$ that are semantically similar to the test input.
The motivation being that observing demonstrations that are similar to the new input text will act as a hint for the model \cite{margatina2023active}.
This requires the use of a retriever $R$, either an off-the-shelf one such as \cite{liu-etal-2022-makes, agrawal-etal-2023-context, margatina2023active, DBLP:journals/corr/abs-2305-14128} or a retriever trained specifically for that task \cite{das-etal-2021-case, rubin-etal-2022-learning}.
For each test example $x_t$, the retriever $R$ is used to rank examples from $\mathcal{D}_{train}$ based on semantic similarity of the text inputs.
Top-$k$ input-output pairs are then selected from the ranked $\mathcal{D}_{train}$ to be used as ICL demonstrations.

\paragraph{Identify Ambiguous Label-Set}
As we can observe from the confusion matrix in \autoref{fig:cm_L}, the model is often confused between two labels.
We hypothesize that in addition to semantic similarity, providing demonstrations that help the model resolve this ambiguity will help the model correct itself. 
Thus, as a next step, we construct a prompt $\theta$ for the test example $x_t$, and use the model log-likelihood to score each output label $l \in L$ given the prompt.
Using this we identify top-2 labels that have the highest scores, which we refer to as the ``\textbf{\ambigls}'' of $x_t$, denoted as $\ambiglst = \{\hat{y}_{t}^{(1)}, \hat{y}_{t}^{(2)}\}$, where $\hat{y}_{t}^{(1)}$ and $\hat{y}_{t}^{(2)}$ are the first and second most likely labels, respectively.

\paragraph{Extract Mis-classified Demonstrations} 
The final component in our recipe is to consider the model prediction of the training data.
While prior work \citet{min-etal-2022-rethinking, yoo-etal-2022-ground, margatina2023active} has looked at training data label-space from the lens of ground-truth labels, i.e., whether to retain them in the ICL or not, we aim to look at label-space from the perspective of model predictions.
Specifically, we are interested in identifying ``hard'' demonstrations, i.e., examples on which the model makes mistakes.
We hope that by showing the model such examples with their ground truth labels will force the model to correct itself.
Prior work has underscored the potential value of leveraging mis-classified examples from the training set to enhance model performance \citep{DBLP:journals/eswa/Tan06, DBLP:conf/aaai/WangZWFSM20}, but they haven't tested it for ICL demonstration selection on text classification.
In addition to the mis-classified examples, we further constrain the model prediction of these mis-classified examples to be one of the ambiguous labels, identified in the above step.
Given that we already know which output labels the model is confused between for the test examples, showing the model those demonstrations (with their ground truth labels) which fall on the decision boundary will likely guide the model to choose the correct label for the test input.

\section{Experimental Setup}
\label{sec:setup}
\subsection{Model}
\label{sec:model}
We experiment with the Flan-PaLM 2 model, an instruction-tuned model which is finetuned on the Flan dataset \citep{DBLP:journals/corr/abs-2210-11416, DBLP:conf/icml/LongpreHVWCTZLZ23} based on PaLM-2  \cite{DBLP:journals/corr/abs-2305-10403}, a multilingual large language model pretrained on web documents, books, code, mathematics and conversational data.
We chose these models as \citealp{DBLP:journals/corr/abs-2305-14128} find that retrieved demonstration for ICL works better with instruction-tuned models over general LLMs (e.g., GPT).
In particular, we experiment with two variants of the model, namely Flan-PaLM-2 (M) and Flan-PaLM-2 (L), where the latter is a larger parameter model.\footnote{Please refer to \citet{DBLP:journals/corr/abs-2305-10403} for more details on the models.}
The ICL demonstrations are selected using an off-the-shelf retriever which is finetuned on mT5-base \citep{xue-etal-2021-mt5} using the unsupervised objective proposed by \citet{DBLP:journals/corr/abs-2112-09118}.
Since the order of demonstrations may impact the model performance \cite{kumar-talukdar-2021-reordering, lu-etal-2022-fantastically}, we randomly shuffle the order of demonstrations for three random seeds and report the average results. 

\subsection{Data} As mentioned above, the Flan-PaLM 2 models are finetuned on the Flan dataset which is a mixture of many supervised datasets.
Specifically, we choose three text classification datasets that satisfy the following desiderata, 1) the output label space shows fine-grained nuance that spans multiple labels, and 2) these datasets are \emph{not} part of the Flan mixture to avoid any inherent bias from the underlying model.
We describe them below, with dataset statistics shown in ~\autoref{tab:data_stat}. All datasets are in English.

\paragraph{EDOS (Task-B):} The Task B of Explainable Detection of Online Sexism \cite{kirk-etal-2023-semeval},  is a \textit{topic classification} task where the sexist content is classified into four categories, i.e., 1) Threats, plans to harm \& incitement, 2) Derogation, 3) Animosity, and 4) Prejudiced Discussion.  

\paragraph{SST:} The Stanford Sentiment Treebank (SST, \citealp{socher-etal-2013-recursive}) is a 5-way \textit{sentiment classification} dataset for movie reviews with labels: Very Negative, Negative, Neutral, Positive, and Very Positive.

\paragraph{GoEmotions:} The GoEmotions \citep{demszky-etal-2020-goemotions} is a multi-class sentiment classification dataset with ``neutral'' and 27 emotional classes, e.g., ``admiration'' and ``fear'', collected from Reddit comments.
As the label space is very large and given that we have limited sequence length, it becomes even more crucial to select a concise but effective prompt.
\footnote{We exclude 24,848 examples (19,925 from training set, 2,474 and 2,449 from dev and test set, respectively) that have multiple labels annotated for a single input, for a simpler experimental setting.
We refer the reader to \citet{demszky-etal-2020-goemotions} for more information on the single-label setting.}

\begin{table}[!t]
\centering
\begin{tabular}{p{.29\linewidth}rrr}\toprule
 & train & dev & test \\\midrule
EDOS & 3,398 & 486 & 970 \\
SST & 8,544 & 1,101 & 2,210 \\
GoEmotions & 23,485 & 2,952 & 2,978 \\\bottomrule
\end{tabular}
\caption{\label{tab:data_stat}Number of examples in each dataset split.  }
\end{table}

\subsection{Baselines}
We compare our proposed method against the following baselines:

\paragraph{Frequent Label (\randguess).} Select the most frequent label as the model prediction for all test examples.

\paragraph{Zero-shot ICL (\base).}\label{zero-shot}
For each test example $x_t$, we prepend the task definition to each test input and prompt the models.\footnote{Please refer to Appendix~\ref{sec:prompt_construction} for the exact prompt and prompt template used in this setting, as well as for few shot settings such as the subsequent \staticn and \topretr.}
To obtain the model prediction, we use the model log-likelihood to score each output label $l \in L$, given the prompt.
Then, we select the label with the highest score.
$y_t = \argmax_{L} \text{score}(l, \theta)$ where $\theta$ refers to the prompt specifically used for this setting, and \emph{score} refers to the model's log-likelihood.

\paragraph{Static N-shot ICL (\staticn).}
We manually select $N$ demonstrations from $\mathcal{D}_{train}$, one for each of the $N$ output labels ($N = |\mathcal{L}|$). 
Note that these demonstrations are static for all test examples.
Thus, we concatenate the task definition, $N$ demonstrations and test example $x_t$  as the prompt for ICL and use the log-likelihood scores, as described above, to get the model prediction. 

\begin{table*}[!t]
\centering
\small
\begin{tabular}{p{.07\linewidth}p{.15\linewidth}p{.07\linewidth}p{.07\linewidth}|p{.07\linewidth}p{.07\linewidth}|p{.07\linewidth}p{.07\linewidth}|p{.04\linewidth}p{.04\linewidth}}\toprule
  & & \multicolumn{2}{c|}{EDOS} & \multicolumn{2}{c|}{SST} & \multicolumn{2}{c|}{GoEmotions} & \multicolumn{2}{c}{Avg.}\\
&  & M & L & M & L & M & L & M & L\\\midrule
\multirow{5}{*}{Baselines} & \randguess & 15.9 & 15.9 & ~~7.5 & ~~7.5 & ~~0.8 & ~~0.8 & ~~8.1 & ~~8.1\\
& \base & 50.7 & 60.5 & 49.2 & 54.1 & \underline{40.5} & 43.4 & \underline{46.8} & 52.7\\
& \staticn & \underline{51.1\std{0.3}} & 58.5\std{0.4} & 50.3\std{0.4}  & \underline{56.5\std{0.3}} & 34.3\std{0.5}  & 44.4\std{0.3} & 45.2& 53.1\\
& \topretr-4 & 48.5\std{0.3}  & \underline{62.3\std{0.4}} & 49.9\std{0.3}  & 55.4\std{0.3} & 38.3\std{0.3}  & 46.2\std{0.4} & 45.6& \underline{54.6}\\
& \topretr-8 & 47.1\std{0.2}  & 61.8\std{0.1} & \underline{51.5\std{0.1}}  & 55.2\std{0.4} & 37.5\std{0.2}  & \underline{46.7\std{0.1}} & 45.4& 54.6\\\midrule
\multirow{8}{*}{Ours} & \textsc{Ambig}-4\\
& \hspace{.5em}{\scriptsize \textsc{+gold}} & 49.3\std{0.6}  & 62.6\std{0.2} & 51.5\std{0.4}  & 56.1\std{0.0} & 40.7\std{0.3}  & \textbf{48.2\std{0.2}} & 47.2& 55.6\\
& \hspace{.5em}{\scriptsize \textsc{+gold+mis}} & 52.2\std{0.5} & 61.7\std{0.9} & 52.3\std{0.1} & 57.4\std{0.1} & 40.1\std{0.2}  & 47.6\std{0.1} & 48.2& 55.6\\
& \hspace{.5em}{\scriptsize \textsc{+gold+mis+pred}} & \textbf{53.9\std{0.5}}  & 62.9\std{0.4} & 53.3\std{0.4}  & \textbf{58.0\std{0.0}} & 42.3\std{0.5}  & 47.7\std{0.2} & \textbf{49.8}& \textbf{56.2}\\
& \textsc{Ambig}-8\\
& \hspace{.5em}{\scriptsize \textsc{+gold}} & 47.5\std{0.1}  & \textbf{63.2\std{0.2}} & 52.9\std{0.1}  & 56.5\std{0.6} & 42.0\std{1.2}  & 47.7\std{0.1} & 47.5& 55.8\\
& \hspace{.5em}{\scriptsize \textsc{+gold+mis}} & 50.4\std{0.4}  & 62.0\std{0.4} & 53.4\std{0.1}  & 57.7\std{0.1} & \textbf{43.9\std{0.2}}  & 47.6\std{0.4} & 49.2& 55.8\\
& \hspace{.5em}{\scriptsize \textsc{+gold+mis+pred}} & 50.9\std{0.6}  & 62.7\std{0.2} & \textbf{54.3\std{0.2}}  & 57.2\std{0.3} & 41.3\std{0.3}  & 47.4\std{0.3} & 48.8& 55.8\\
\bottomrule
\end{tabular}
\caption{\label{result_ML} F1 macro (\%) comparison between our baselines (top) and our proposed methods (bottom) with Flan-PaLM 2 (M/L). 4 or 8 refers to the number of ICL demonstrations.
The best performance across all method is \textbf{highlighted} and the best performing baseline is \underline{underlined}. The ``Avg.'' column shows the average scores across all datasets. The standard deviations are computed over three random seeds, with the order of demonstrations shuffled. 
}
\end{table*}

\begin{table}[!t]
\centering
\small
\begin{tabular}{lll|lll}\toprule
\multirow{2}{*}{} & \multirow{2}{*}{\base} & \multirow{2}{*}{\staticn} &  \multicolumn{3}{c}{\textsc{Ambig-ICL}} \tnote{a}\\
 &  &  & \textsc{+gold} & \textsc{+mis} & \textsc{+pred} \\\midrule
M & ~1.3 & -0.2 & 1.9 & 3.3 & \textbf{3.9} \\
L & -1.9 & -1.5 & 1.1 & 1.1 & \textbf{1.4} \\\midrule
all & -0.3 & -0.9 & 1.5 & 2.2 & \textbf{2.6} \\\bottomrule
\end{tabular}
\begin{tablenotes}
{\scriptsize
\item[a] We omitted \topretr in the table, which are inherently zero as we compare against \topretr.
\item[b] For both \topretr and \textsc{Ambig-ICL}, we average results on both 4 and 8 shots before computing differences.
}
\end{tablenotes}
\caption{\label{tab:f1_diff} F1 macro (\%) differences compared to \topretr, averaged across all datasets as detailed in \autoref{result_ML}. M and L refers to Flan-PaLM 2 sizes, and ``all'' is averaged on results of size M and L. ``\textsc{+mis}'' and ``\textsc{+pred}'' refer to ``\textsc{+gold+mis}'' and ``\textsc{+gold+mis+pred}'', respectively.
}
\end{table}

\paragraph{Retriever-based ICL (\topretr).}
Unlike above, where we used the same prompt for all test inputs, in this baseline, we retrieve demonstrations for each test input $x_t$.
We use an off-the-shelf retriever $R$ (\autoref{sec:model}) to retrieve $k$ nearest neighbors $\{x_{1,t}, \cdots, x_{k,t}\}$ from $\mathcal{D}_{train}$, similar to \citet{das-etal-2021-case}. We encode the input text of training set and the test example, rank the training data by the inner product of the vectors.  
Of these $k$ examples, we select $n=4,8$ as ICL demonstrations.\footnote{
We chose $k = 4, 8$ for two reasons: a) to limit the sequence length to 1024 tokens for faster inference, and b) in some settings we found $k = 4$ often outperforming $k = 8$ (\autoref{result_ML}), which led us to believe that adding more examples will not benefit much.
}

\subsection{Proposed Method: \textsc{Ambig}-ICL}
As described in \autoref{sec:method}, our proposed method considers both semantic similarity and the label ambiguity for selecting demonstrations.
Below, we summarize our proposed model variants.
For each setting, we first retrieve the top-$k$ most similar examples  from the training data $\mathcal{D}_{train}$ for each test example $x_t$.
We denote these candidates by  $R(x_t)= \{(x_{0,t}, y_{0,t}), \cdots, (x_{k,t}, y_{k,t})\}$.
At the same time, for each $x_t$, we also identify the ambiguous label-set 
$\ambiglst = \{l_i, l_j | l \in L\}$. This set contains the top-2 labels, $l_i$ and $l_j$, that the model is most confused about, where both labels belong to the set $L$ of all output labels.

\paragraph{+\textsc{gold}}
Select those examples from $R(x_t)$ as demonstrations where the ground truth label of each demonstration belongs to the ambiguous label set of $x_t$ denoted by:
\[\textsc{icl} (x_t) = \left\{
    \begin{array}{ll}
    (x_i, y_i) \text{ if } y_i \in \ambiglst \\
    \text{for } (x_i, y_i) \in R(x_t) 
    \end{array} \right \}\]

\paragraph{+\textsc{gold}+\textsc{mis}}
Select those examples from $R(x_t)$ as demonstrations where the ground truth labels fall in $\ambiglst$ and they are mis-classified, denoted by:
\[ \textsc{icl} (x_t) = \left \{
\begin{array}{ll}
     (x_i, y_i) \text{ if } y_i \in \ambiglst, \hat{y}_i \neq y_i  \\
     \text{for } (x_i, y_i) \in R(x_t)
\end{array} \right \} \]
Note that the model predictions ($\hat{y}$) on the $R(x_t)$ are obtained from the \base model.    

\paragraph{+\textsc{gold}+\textsc{mis}+\textsc{pred}}
Select those examples from $R(x_t)$ as demonstrations where the ground truth labels fall in $\ambiglst$. Ensure they are mis-classified and with an additional constraint, that their model predictions also fall within $\ambiglst$, denoted by:
\[\textsc{icl} (x_t) = \left\{
    \begin{array}{ll}
    (x_i, y_i) \text{ if } y_i \in \ambiglst, \hat{y}_i \neq y_i,  \\
    \hat{y}_i \in \ambiglst \text{ for } (x_i, y_i) \in R(x_t) 
    \end{array} \right \}\]
Same as above, the model predictions on the training data are obtained from \base.

For all our proposed model variants, we select $n$ demonstrations where $n=4$ and $n=8$. 

\begin{table*}[!t]
    \centering
    \small
    \begin{tabular}{p{0.14\linewidth}p{0.65\linewidth}p{0.1\linewidth}}\toprule
    \multicolumn{3}{l}{\textbf{Test Example: Ok! I like making friends} \hfill 
    \textbf{$\mathbf{\ambiglst}$: Love, Joy} \hfill \textbf{Gold label: Love}} \\\midrule
    \topretr & \begin{minipage}[t]{\linewidth}
    1. Disappointment: I want to \textit{make friends} too :( but I feel like I have nothing good to offer\\
    2. Joy: I, too, am a lot of fun at parties. We can stand together in the corner! \\
    3. Gratitude: Thanks. I am. I \textit{make some new friends}.\\
    4. Disapproval: Not really. My group of \textit{friends} are awesome in every way possible except they are homophobic
    \end{minipage} & Predicted: Joy \\\midrule
    \multicolumn{3}{l}{\textsc{Ambig-ICL}}  \\
    \hspace{.5em}{\scriptsize\textsc{+gold}} & \begin{minipage}[t]{\linewidth}
    1. Joy: I, too, am a lot of fun at parties. We can stand together in the corner! \\
    2. Love: I ... I like you\\
    3. Love: Married to the love of my life. LOL\\
    4. Love: I do. but some people love it\\
    \end{minipage} & Predicted: Love\\
    \hspace{.5em}{\scriptsize \textsc{+gold+mis}}& \begin{minipage}[t]{\linewidth}
    1. Joy: I, too, am a lot of fun at parties. We can stand together in the corner! \\
    2. Love: Too cute for me. Why cant i have a boyfriend *[NAME]* \\
    3. Joy: FaceTime with wifey!! Happy anniversary! \\
    4. Love: Stick around! Would love your input \/ POV! \\
    \end{minipage} & Predicted: Love\\
    \hspace{.5em}{\scriptsize \textsc{+gold+mis+pred}} & \begin{minipage}[t]{\linewidth}
    1. Joy: FaceTime with wifey!! Happy anniversary!\\
    2. Joy: She want to take it slow, I can see that... I deal with those girls all the time, they my favorite\\
    3. Love: Ha! I like that one.\\
    4. Love: Ooh I like that one :)
    \end{minipage}  & Predicted: Love\\
    \bottomrule
    \end{tabular}
    \caption{\label{tab:icl_comparison} Example demonstrations selected by the \topretr and our proposed method \textsc{Ambig-ICL} for the GoEmotions task, for $n=4$.
    Each demonstration comprises of the input text and the ground truth label, as selected from the training data.
    On Flan-PaLM 2 (L), where \topretr mis-classified it as ``Joy'', \textsc{Ambig-ICL} predicted correctly under all three settings. 
    }
\end{table*}

\section{Results and Discussion}
We report all our results in \autoref{result_ML}. Specifically, we use the F1 macro scores to compare the model performance, as all our tasks have unbalanced datasets.\footnote{We report the accuracy, precision and recall in \ref{sec:other_results}.}
First, we note across all three tasks, our proposed methods outperform the baselines.

We also note that the zero-shot model (\base) which only uses a task definition but no task demonstrations, already is a strong baseline for both the Flan-PaLM 2 models (M/L). In particular, comparing the average scores of the few-shot baselines and \base, we find that \base outperforms few-shot baselines by 1.4\% on Flan-PaLM 2 (M), but the larger model Flan-PaLM 2 (L) benefits from the addition of ICL demonstrations (+1.4\% gain).
This is because larger-parameter models make better use of in-context learning \citep{DBLP:journals/corr/abs-2210-05675, DBLP:conf/iclr/AkyurekSA0Z23, DBLP:journals/corr/abs-2303-03846}.
Interestingly, we also observe that for SST and GoEmotions, the Flan-PaLM 2 (L) model achieves higher performance with $n=4$ over $n=8$, which highlights that quantity does not necessarily lead to better performance.

\paragraph{Considering output label space is more important than semantic similarity.}
Within the few-shot methods, where we use ICL demonstrations along with the task definition, we compute from \autoref{tab:f1_diff} that our proposed methods \textsc{ambig-*} outperforms retriever-based models (\topretr-*) by  +3.0\% (avg.) for Flan-PaLM 2 (M), and by +1.2\% (avg.) for Flan-PaLM 2 (L), suggesting that \emph{considering output label space for selecting demonstrations is as important as considering the input similarity.}
In particular, we find that considering mis-classified demonstrations that fall on the test example's decision boundary leads to the overall best performance.
In ~\autoref{tab:icl_comparison}, we show the demonstrations selected for the $n=4$ setting for one example of the GoEmotions task. 
We see that for the test input ``Ok! I like making friends'', the \topretr method retrieved similar examples from $\mathcal{D}_{train}$ (all examples refer to \emph{friends}).
Now from the \base model, we calculated the model prediction scores and found that \emph{Love} and \emph{Joy} are the two labels the model is most confused about.
However, because we do not consider any test example ambiguity in \topretr, only one of the retrieved examples represent the labels \emph{Love} or \emph{Joy}, which are the two labels the model is most confused about for this test example.
Whereas, in the \textsc{Ambig-ICL} setting, because of our constraints, all the examples chosen for ICL belong to the ambiguous label set.
This allows all our proposed methods to better understand this fine-grained nuance across label space and make the correct model prediction of \emph{Love}.
Below, we conduct some analysis to further explain the way our proposed methods work.

\paragraph{Considering output label space compensates for the sacrifice in semantic similarity.}
As we introduce more constraints (i.e., \textsc{+gold}, \textsc{+mis}, and \textsc{+pred}), we find that we need to sacrifice the semantic similarity to the test input.
For example, consider the 4-shot \textsc{Ambig-ICL} experiment on EDOS (Task-B), to satisfy the constraints for the \textsc{+gold} setting we need to select up to top-16 retrieved examples in order to obtain the 4 ICL demonstrations; for \textsc{+gold+mis} we need top-55 retrieved examples and more than top-250 retrieved examples for \textsc{+gold+mis+pred}.\footnote{We set a strict constraint on our selection (top-250 retrieved example for \textsc{+gold}, and top-250 misclassified retrieved examples for the other two). If there aren't sufficient examples for \textsc{+gold+mis+pred} within the top-250 misclassified retrieved example, we fall-back on the previous setting (\textsc{+gold+mis}).
}
Clearly, by selecting lower ranked examples from the retrieved set $R(x_t)$ we are sacrificing the semantic similarity to the test input.
While previous studies, such as \citep{das-etal-2021-case,liu-etal-2022-makes, margatina2023active}, have indicated that greater semantic similarity can enhance model performance, we can see that our methods can still outperform the retriever-based baselines which prioritize it.

\paragraph{The ambiguous label set is a good proxy for the test gold label.}
While \citet{min-etal-2022-rethinking} find that using pseudo-demonstrations i.e. demonstrations with random labels instead of the ground truth labels, does not affect the downstream performance much, \citet{lyu-etal-2023-z} find that for demonstrations that are similar to the test input, such as those from a retriever, pseudo-demonstrations hurt the performance. 
They refer to this as the copying-effect hypothesis which says that the ``model prediction is biased towards the labels paired with the inputs in the demonstrations, especially when the inputs are similar to the test inputs''.
This, in turn, suggests that the best performance could be achieved if the labels paired with the inputs are same as the gold label of the test example.
Given that we do not know the gold label of the test example apriori, the question then becomes \emph{how do we approximate the gold label?}.
We find that our \emph{ambiguous label set} acts as a close proxy.
In \autoref{gold_label_num}, we compute how many times is the label paired with ICL demonstrations the same as the test example gold label. 
We find that 44.2\% of our proposed methods' (\textsc{ambig}) demonstrations have the same gold label as the test example on average, compared to 30.9\% from the \topretr method.
This is why including the ambiguous label set in the demonstration selection process leads to a higher performance.
This analysis also sheds light on the effectiveness of retriever-based ICL. 
From \autoref{gold_label_num} we can see that the demonstrations selected solely based on input text similarity is only 13.3\% points (avg.) behind our proposed methods.
This confirms that finding demonstrations similar to the input text also leads to selecting demonstrations that have the `likely' gold label.

\begin{table}[!t]
\centering
\small
\begin{tabular}{p{0.28\linewidth}p{0.04\linewidth}p{0.06\linewidth}|p{0.04\linewidth}p{0.06\linewidth}|p{0.05\linewidth}p{0.07\linewidth}}\toprule
\multirow{2}{*}{} & \multicolumn{2}{c|}{EDOS} & \multicolumn{2}{c|}{SST} & \multicolumn{2}{c}{GoEmotions} \\
 & M & L & M & L & M & L \\\midrule
$4$-shot & \multicolumn{2}{c|}{42.6} & \multicolumn{2}{c|}{29.6} & \multicolumn{2}{c}{21.6} \\
$8$-shot & \multicolumn{2}{c|}{42.5} & \multicolumn{2}{c|}{28.6} & \multicolumn{2}{c}{20.5} \\\midrule
\textsc{Ambig}-4\\
\hspace{.5em}{\scriptsize \textsc{+gold}} & 49.5 & \textbf{50.3} & \textbf{46.5} & \textbf{47.1} & \textbf{41.3} & \textbf{41.9} \\
\hspace{.5em}{\scriptsize \textsc{+gold+mis}} & 46.4 & 44.3 & 46.1 & 44.3 & 38.7 & 38.8 \\
\hspace{.5em}{\scriptsize \textsc{+gold+mis+pred}} & 48.3 & 42.3 & 46.1 & 44.6 & 37.8 & 40.7 \\
\textsc{Ambig}-8\\
\hspace{.5em}{\scriptsize \textsc{+gold}} & \textbf{50.3} & \textbf{50.3} & 46.0 & 46.8 & 41.2 & 41.7 \\
\hspace{.5em}{\scriptsize \textsc{+gold+mis}} & 46.9 & 43.8 & 46.4 & 44.7 & 38.7 & 38.6 \\
\hspace{.5em}{\scriptsize \textsc{+gold+mis+pred}} & 48.8 & 42.9 & \textbf{46.5} & 44.9 & 37.5 & 40.3\\\bottomrule
\end{tabular}
\caption{\label{gold_label_num} Average percentage (\%) of examples in the top $4, 8$ retrieved demonstrations that share the same gold labels with test example. 
}
\end{table}

\begin{table}[!t]
\centering
\small
\begin{tabular}{p{0.28\linewidth}p{0.04\linewidth}p{0.06\linewidth}|p{0.04\linewidth}p{0.06\linewidth}|p{0.05\linewidth}p{0.07\linewidth}}\toprule
\multirow{2}{*}{} & \multicolumn{2}{c|}{EDOS} & \multicolumn{2}{c|}{SST} & \multicolumn{2}{c}{GoEmotions} \\
 & M & L & M & L & M & L \\\midrule
 uniform & \multicolumn{2}{c|}{2.00} & \multicolumn{2}{c|}{2.32}& \multicolumn{2}{c}{4.75}\\\midrule
\base & 0.98 & 1.08 & 1.58 & 1.19 & 2.44 & 1.92 \\
\staticn & 0.87 & 1.07 & 1.41 & 1.11 & 1.76 & 1.77 \\
\topretr-$4$ & 0.78 & 0.97 & 1.40 & 1.06 & 1.89 & 1.70 \\
\topretr-$8$ & 0.82 & 0.96 & 1.38 & 1.04 & 1.79 & 1.69 \\\midrule
\textsc{Ambig}-4\\
\hspace{.5em}{\scriptsize \textsc{+gold}} & \textbf{0.77} & 0.93 & 1.39 & 1.02 & 1.86 & 1.43 \\
\hspace{.5em}{\scriptsize \textsc{+gold+mis}} & 0.85 & 0.98 & 1.41 & 1.06 & 1.92 & 1.48 \\
\hspace{.5em}{\scriptsize \textsc{+gold+mis+pred}} & 0.86 & 1.00 & 1.42 & 1.07 & 1.92 & 1.46 \\
\textsc{Ambig}-8\\
\hspace{.5em}{\scriptsize \textsc{+gold}} & 0.81 & \textbf{0.91} & \textbf{1.36} & \textbf{0.98} & \textbf{1.68} & \textbf{1.33} \\
\hspace{.5em}{\scriptsize \textsc{+gold+mis}} & 0.89 & 0.97 & 1.39 & 1.03 & 1.74 & 1.39 \\
\hspace{.5em}{\scriptsize \textsc{+gold+mis+pred}} & 0.90 & 1.00 & 1.40 & 1.04 & 1.76 & 1.37\\
\bottomrule
\end{tabular}
\caption{\label{tab:entropy} Average entropy of predicted probability distribution. ``uniform'' refers to the entropy computed for an uniform probability distribution over the labels. Lower entropy is better.
}
\end{table}

\paragraph{\textsc{Ambig-ICL} helps reduce the model confusion.}
To understand whether including test label ambiguity indeed helps decrease the model confusion, we calculate the model entropy over the predicted probability distribution of the output labels in \autoref{tab:entropy}.\footnote{We compute entropy with a base of 2.}
Overall, we observe that our \textsc{Ambig-*} methods achieve the lowest entropy across all three datasets and models.
This suggests that by explicitly identifying the point of model confusion (in this case the confusion across fine-grained labels) and selecting demonstrations that help resolve this confusion is indeed effective in reducing the confusion across labels, and thereby resulting in higher downstream performance (\autoref{result_ML}). 
In particular, we find that for the Flan-PaLM 2 (L), the gap between the few-shot baselines and the \textsc{Ambig-*} methods is larger, perhaps because larger models are better able to use the ICL demonstrations  \citep{DBLP:journals/corr/abs-2210-05675, DBLP:conf/iclr/AkyurekSA0Z23, DBLP:journals/corr/abs-2303-03846}.

We also compute the Pearson correlation coefficient between F1 macro scores and average entropy of predicted probability distribution (shown in \autoref{result_ML} and \autoref{tab:entropy}, respectively), for all the three datasets. 
We find that for the Flan-PaLM 2 (L) model, there is a negative correlation for all three datasets, i.e., $r\!=\!-0.78$ for EDOS, $-0.48$ for SST and $-0.92$ for GoEmotions, which suggests that lower entropy translates to higher task performance.
However, for the Flan-PaLM 2 (M), we have mixed results, as $r$ is positive for EDOS ($0.47$), negative for SST ($-0.55$), and close to zero for GoEmotions ($0.03$).

\section{Related Work}

The performance of large language models (LLMs) is significantly influenced by the quality of ICL demonstrations, as demonstrated in multiple studies \citep{DBLP:conf/icml/ZhaoWFK021, liu-etal-2022-makes, zhang-etal-2022-active}. 
Consequently, the focus on retrieving superior demonstrations has increased.
One prominent strategy is to finetune a retriever for specific tasks by similarity metrics \citep{das-etal-2021-case, hu-etal-2022-context, DBLP:conf/iclr/PoesiaP00SMG22} or by scores derived from language models \citep{rubin-etal-2022-learning, shi-etal-2022-xricl}. 
While some works introduce an unified retriever trained across various tasks \citep{li-etal-2023-unified, DBLP:journals/corr/abs-2303-08518} for generalizabilty, another direction is to leverage off-the-shelf retrievers. 
\citealp{liu-etal-2022-makes} propose a KNN-based method to select ICL demonstrations based on semantic similarities;  \citealp{margatina2023active}  select ICL demonstrations with active learning algorithms based on uncertainty, diversity, and similarity, and show that selecting based on input text similarity consistently outperforms other methods; and \citealp{agrawal-etal-2023-context} focus on selecting diverse demonstrations as well as promoting n-gram overlap between demonstrations and test examples.
In our work, we adopt the off-the-shelf retriever approach as our focus is to show the generalizability of our approach across different classification tasks.  However, we expect that our method will also benefit from a task-specific  retriever.
Additionally, to the best of our knowledge, we are the first ones to leverage the LLM's existing knowledge surrounding the test example for selecting demonstrations.
Prior works have typically explored the LLM's existing knowledge, considering the model prediction for the training data.

 \citealp{DBLP:journals/corr/abs-2305-14128} use the LLM prediction score on the training data to train a task-specific retriever, and also use Chain-of-Thought prompting \citep{DBLP:conf/nips/Wei0SBIXCLZ22} to improve model performance.
Some works \citep{kumar-talukdar-2021-reordering, lu-etal-2022-fantastically} have found that ordering of the ICL demonstrations also affects the downstream performance, that is why in \autoref{result_ML} we report the results across three shuffle orders. 
These works  are orthogonal to our work but can be used in combination with our proposed methods.

\section{Conclusion and Next Steps}
In this work, we find that using LLM's existing knowledge (e.g., the model prediction) regarding the output label space of both the test example and the ICL demonstration pool is as important as considering the semantic similarity of the input text alone.
We find that our proposed method consistently outperform the baselines for all three tasks.
Although, we only consider the top-2 most ambiguous labels in selecting the ICL demonstrations, it would be interesting to expand the ambiguous label set to more than two labels.
This would especially be more important for datasets like GoEmotions where the label space is large and much more fine-grained.
We leave this effort for future work.
Furthermore, in this work, we focus on sentence classification tasks, thus paving the way for others  to use our proven techniques to also explore label ambiguity for other token/span-level tasks such as Named Entity Recognition (NER), and Part-Of-Speech (POS) tagging.

\section{Limitations}
We focus on reducing LLM's label ambiguity by incorporating demonstrations that are misclassified by the LLM and reside on the test example's decision boundary. While we show this methodology's effectiveness across datasets, even those with a granular label structure, potential pitfalls remain. If the actual gold label of test example often deviates from the LLM's top two label choices in a particular dataset or model, this can be indicative of subpar zero-shot performance or flawed ambiguous label set selection. In these scenarios, our method may lead to unsatisfying performance, necessitating further enhancements.

\section{Ethics Statement}
We use pretrained large language models (LLMs) for text classification. Notably, LLMs are shown to exhibit biases, which is a well-recognized challenge and the broader community is currently working to address. 
Since our main goal is to improve the downstream task performance, an improved performance on an offensive content classification task could be misused.
In particular, the EDOS dataset used in our work, contains offensive content. 
We selected this dataset for its fine-grained label nuances and to ensure our research isn't biased by models inherently familiar with the data. 

\bibliography{anthology,custom}
\bibliographystyle{acl_natbib}
\clearpage
\appendix

\section{Appendix}\label{sec:appendix}

\subsection{Prompt Construction}\label{sec:prompt_construction}

We show our templates in \autoref{tab:prompt_template} (we use 4-shot as an example for few-shot). Task definitions are listed below, denoted by $x_{defn}$: 

{\small
\begin{itemizesquish}
\item \textbf{EDOS:} Given a text input, the task is to classify the input as being a Threat, Prejudiced, Animosity, or Derogation category of sexism. Threat refers to language where an individual expresses intent and\/or encourages others to take action against women which inflicts or incites serious harm and violence against them. It includes threats of physical, sexual or privacy harm. Prejudiced refers to language which denies the existence of discrimination, and justifies sexist treatment. It includes denial and justification of gender inequality, excusing women's mistreatment, and the ideology of male victimhood. Animosity refers to language which expresses implicit or subtle sexism, stereotypes or descriptive statements. It includes benevolent sexism, i.e., framed as a compliment. Derogation refers to language which explicitly derogates, dehumanises, demeans or insults women. It includes negative descriptions and stereotypes about women, objectification of their bodies, strong negative emotive statements, and dehumanising comparisons. It covers negative statements directed at a specific woman and women in general.
\item \textbf{SST:} Given sentences from movie reviews, the task is to classify the sentences as being a Great, Good, Okay, Bad, or Terrible category of sentiment. Great refers to language that expresses extremely positive sentiment. Good refers to language that expresses positive sentiment, but not to the extreme. Okay refers to language that is neutral, i.e., neither expresses clear positive nor negative sentiments. Bad refers to language that expresses negative sentiment, but not to the extreme. Terrible refers to language that expresses extremely negative sentiment.
\item \textbf{GoEmotions:} Given sentences from Reddit comments, the task is to classify the sentences as being an Admiration, Approval, Annoyance, Gratitude, Disapproval, Amusement, Curiosity, Love, Optimism, Disappointment, Joy, Realization, Anger, Sadness, Confusion, Caring, Excitement, Surprise, Disgust, Desire, Fear, Remorse, Embarrassment, Nervousness, Pride, Relief, or Grief category of emotions.
\end{itemizesquish}
}

\begin{table}[]
    \centering
    \small
    \noindent\fbox{%
    \begin{minipage}{\dimexpr\linewidth-2\fboxsep-2\fboxrule} 
\tt
$x_{defn}$\\
\\
Thus given the following input:\\
input: $x_t$\\
answer:
    \end{minipage}
}
\noindent\fbox{%
    \begin{minipage}{\dimexpr\linewidth-2\fboxsep-2\fboxrule} 
\tt
$x_{defn}$\\
\\
Some examples are:\\
input: $x_{1, t}$\\
answer: $y_{1, t}$\\
\\
input: $x_{2, t}$\\
answer: $y_{2, t}$\\
\\
input: $x_{3, t}$\\
answer: $y_{3, t}$\\
\\
input: $x_{4, t}$\\
answer: $y_{4, t}$\\
\\
Thus given the following input:\\
input: $x_t$\\
answer: 
    \end{minipage}
}
\caption{Prompt templates for zero-shot and few-shot ICL. $x_t$ refers to the test example, and $x_{i,t}, y_{i,t}$ refers to the text inputs and gold labels of ICL demonstrations selected for $x_t$, respectively.}
\label{tab:prompt_template}
\end{table}

\subsection{Accuracy, Precision, Recall}\label{sec:other_results}

Please refer to \autoref{tab:prec}, \ref{tab:recall} and \ref{tab:acc}.

\begin{table*}[!t]
\small
\centering
\begin{tabular}{llllllll}\toprule
\multicolumn{2}{c}{\multirow{2}{*}{}} & \multicolumn{2}{c}{EDOS} & \multicolumn{2}{c}{SST} & \multicolumn{2}{c}{GoEmotions} \\
\multicolumn{2}{c}{} & M & L & M & L & M & L \\\midrule
\multirow{5}{*}{Baselines} & \randguess & 11.7 & 11.7 & ~~4.6 & ~~4.6 & ~~0.4 & ~~0.4 \\
 & \base & 65 & 60.7 & 54 & 56.2 & 42.6 & 46.3 \\
 & \staticn & 65.2\std{0.6} & 58.1\std{0.4} & 54.5\std{0.6} & 58.2\std{0.3} & 42.6\std{1.2} & 46.2\std{0.3} \\
 & \topretr-4 & 67.1\std{1.1} & 63.6\std{0.5} & 53.4\std{0.3} & 57.4\std{0.4} & 43.7\std{0.4} & 47.6\std{0.4} \\
 & \topretr-8 & 65.0\std{0.2} & 63.9\std{0.3} & 54.4\std{0.1} & 57.6\std{0.5} & 43.7\std{0.4} & 48.3\std{0.1} \\\midrule
\multirow{8}{*}{Ours} & \textsc{Ambig}-4\\
 & \hspace{.5em}{\scriptsize \textsc{+gold}} & 65.9\std{0.8} & 63.6\std{0.4} & 54.1\std{0.3} & 57.7\std{0.1} & 45.7\std{0.3} & 50.5\std{0.2} \\
 & \hspace{.5em}{\scriptsize \textsc{+gold+mis}} & 66.6\std{1.1} & 63.6\std{1.0} & 54.1\std{0.2} & 58.8\std{0.1} & 44.8\std{0.4} & 49.2\std{0.1} \\
 & \hspace{.5em}{\scriptsize \textsc{+gold+mis+pred}} & 67.4\std{0.4} & 65.0\std{0.5} & 54.8\std{0.5} & 59.4\std{0.0} & 46.9\std{1.3} & 47.9\std{0.2} \\
 & \textsc{Ambig}-8\\
 & \hspace{.5em}{\scriptsize \textsc{+gold}} & 66.4\std{1.1} & 64.8\std{0.1} & 54.7\std{0.2} & 58.5\std{0.7} & 48.0\std{1.8} & 49.9\std{0.1} \\
 & \hspace{.5em}{\scriptsize \textsc{+gold+mis}} & 68.4\std{0.8} & 64.4\std{0.6} & 54.5\std{0.1} & 59.6\std{0.1} & 48.7\std{0.5} & 48.8\std{0.5} \\
 & \hspace{.5em}{\scriptsize \textsc{+gold+mis+pred}} & 66.6\std{1.2} & 66.4\std{0.3} & 54.9\std{0.2} & 59.1\std{0.4} & 43.7\std{0.5} & 47.4\std{0.3} \\\bottomrule
\end{tabular}
\caption{\label{tab:prec} Precision (\%) comparison between our proposed methods and baselines with Flan-PaLM 2 (M, L).}
\end{table*}

\begin{table*}[!t]
\small
\centering
\begin{tabular}{llllllll}\toprule
\multicolumn{2}{c}{\multirow{2}{*}{}} & \multicolumn{2}{c}{EDOS} & \multicolumn{2}{c}{SST} & \multicolumn{2}{c}{GoEmotions} \\
\multicolumn{2}{c}{} & M & L & M & L & M & L \\\midrule
\multirow{5}{*}{Baselines} 
 & \randguess & 25 & 25 & 20 & 20 & ~~3.7 & ~~3.7 \\
 & \base & 46 & 62.8 & 53.8 & 55.2 & 42.4 & 47.2 \\
 & \staticn & 46.2\std{0.3} & 63.0\std{0.3} & 54.0\std{0.4} & 56.5\std{0.2} & 34.8\std{0.5} & 49.5\std{0.4} \\
 & \topretr-4 & 44.8\std{0.3} & 63.4\std{0.2} & 53.4\std{0.3} & 55.7\std{0.3} & 38.5\std{0.2} & 49.7\std{0.3} \\
 & \topretr-8 & 44.0\std{0.1} & 62.1\std{0.2} & 54.2\std{0.1} & 55.3\std{0.4} & 37.8\std{0.3} & 50.1\std{0.3} \\\midrule
\multirow{8}{*}{Ours} & \textsc{Ambig}-4\\
 & \hspace{.5em}{\scriptsize \textsc{+gold}} & 45.1\std{0.6} & 64.1\std{0.2} & 54.6\std{0.4} & 56.4\std{0.1} & 41.4\std{0.3} & 51.3\std{0.2} \\
 & \hspace{.5em}{\scriptsize \textsc{+gold+mis}} & 48.0\std{0.4} & 62.1\std{0.9} & 54.9\std{0.1} & 57.3\std{0.1} & 40.9\std{0.1} & 51.0\std{0.4} \\
 & \hspace{.5em}{\scriptsize \textsc{+gold+mis+pred}} & 49.5\std{0.4} & 63.1\std{0.3} & 55.6\std{0.4} & 57.7\std{0.0} & 42.7\std{0.2} & 51.7\std{0.4} \\
 & \textsc{Ambig}-8\\
 & \hspace{.5em}{\scriptsize \textsc{+gold}} & 43.6\std{0.1} & 64.0\std{0.2} & 55.0\std{0.1} & 56.5\std{0.6} & 41.9\std{0.9} & 50.8\std{0.5} \\
 & \hspace{.5em}{\scriptsize \textsc{+gold+mis}} & 47.3\std{0.4} & 61.8\std{0.3} & 54.9\std{0.1} & 57.3\std{0.1} & 44.4\std{0.2} & 51.4\std{0.3} \\
 & \hspace{.5em}{\scriptsize \textsc{+gold+mis+pred}} & 48.0\std{0.4} & 61.7\std{0.2} & 55.6\std{0.2} & 56.7\std{0.2} & 43.2\std{0.3} & 51.3\std{0.1}
 \\\bottomrule
\end{tabular}
\caption{\label{tab:recall} Recall (\%) comparison between our proposed methods and baselines with Flan-PaLM 2 (M, L).}
\end{table*}

\begin{table*}[!t]
\small
\centering
\begin{tabular}{llllllll}\toprule
\multicolumn{2}{c}{\multirow{2}{*}{}} & \multicolumn{2}{c}{EDOS} & \multicolumn{2}{c}{SST} & \multicolumn{2}{c}{GoEmotions} \\
\multicolumn{2}{c}{} & M & L & M & L & M & L \\\midrule
\multirow{5}{*}{Baselines} 
 & \randguess & 46.8 & 46.8 & 23.1 & 23.1 & 11.7 & 11.7 \\
 & \base & 55.4 & 59.2 & 49.9 & 57.1 & 47.1 & 46.2 \\
 & \staticn & 54.3\std{0.3} & 57.6\std{0.1} & 50.5\std{0.4} & 59.0\std{0.2} & 39.8\std{0.2} & 46.7\std{0.2} \\
 & \topretr-4 & 53.6\std{0.3} & 61.0\std{0.5} & 50.0\std{0.3} & 58.5\std{0.3} & 45.9\std{0.0} & 50.1\std{0.2} \\
 & \topretr-8 & 53.8\std{0.3} & 61.1\std{0.2} & 51.8\std{0.1} & 58.6\std{0.4} & 45.3\std{0.0} & 51.0\std{0.2}  \\\midrule
\multirow{8}{*}{Ours} & \textsc{Ambig}-4\\
 & \hspace{.5em}{\scriptsize \textsc{+gold}} & 54.3\std{0.4} & 61.3\std{0.4} & 51.5\std{0.4} & 58.9\std{0.1} & 46.6\std{0.2} & 50.3\std{0.1} \\
 & \hspace{.5em}{\scriptsize \textsc{+gold+mis}} & 56.1\std{0.2} & 60.9\std{0.6} & 52.1\std{0.1} & 59.7\std{0.1} & 45.6\std{0.1} & 49.5\std{0.1} \\
 & \hspace{.5em}{\scriptsize \textsc{+gold+mis+pred}} & 56.5\std{0.1} & 61.4\std{0.4} & 53.0\std{0.4} & 60.1\std{0.1} & 45.6\std{0.1} & 50.0\std{0.2} \\
 & \textsc{Ambig}-8\\
 & \hspace{.5em}{\scriptsize \textsc{+gold}} & 53.6\std{0.2} & 61.8\std{0.0} & 52.9\std{0.1} & 59.5\std{0.6} & 46.8\std{0.1} & 50.4\std{0.2} \\
 & \hspace{.5em}{\scriptsize \textsc{+gold+mis}} & 55.4\std{0.6} & 61.1\std{0.3} & 53.2\std{0.1} & 60.2\std{0.1} & 45.8\std{0.2} & 50.0\std{0.3} \\
 & \hspace{.5em}{\scriptsize \textsc{+gold+mis+pred}} & 55.1\std{0.6} & 61.5\std{0.3} & 54.2\std{0.2} & 59.6\std{0.3} & 44.9\std{0.2} & 50.2\std{0.2}
 \\\bottomrule
\end{tabular}
\caption{\label{tab:acc} Accuracy (\%) comparison between our proposed methods and baselines with Flan-PaLM 2 (M, L).}
\end{table*}

\subsection{Label-wise Percentage Analysis of Gold Label Inclusion in \texorpdfstring{$\mathbf{\ambiglst}$}{L\_\{ambig, t\}}}\label{sec:label-wise-LIR}

We compute the percentage of times that the test example's gold label is in $\ambiglst$ (as obtained with \base) in \autoref{LIR}, and we present label-wise results in \autoref{tab:label-wise-lir}.

\begin{table}[!t]
\centering
\small
\begin{tabular}{p{0.3\linewidth}p{0.14\linewidth}p{0.14\linewidth}p{0.18\linewidth}}\toprule
 & EDOS & SST & GoEmotions \\\midrule
Flan-PaLM 2 (M) & 91.2 & 85.8 & 61.2 \\
Flan-PaLM 2 (L) & 88.2 & 87.6 & 61.6 \\\bottomrule
\end{tabular}
\caption{\label{LIR} Percentage of times the test example's gold label is in $\ambiglst$ (as obtained from \base model).}
\end{table}

\begin{table*}[!t]
\centering
\small
\begin{tabular}{p{0.1\linewidth}p{0.01\linewidth}p{0.8\linewidth}}\toprule
\multirow{2}{*}{EDOS} & M & Animosity 99.1, Derogation 97.4, Prejudiced   52.1, Threat 71.9 \\
 & L & Animosity 90.1, Derogation 90.1, Prejudiced 68.1, Threat 93.3 \\\midrule
\multirow{2}{*}{SST} & M & Bad 78.4, Good 98.0, Great 88.0, Okay 73.8, Terrible 93.9 \\
 & L & Bad 89.3, Good 99.2, Great 99.2, Okay 59.1, Terrible 85.3 \\\midrule
 \multirow{2}{*}{GoEmotions} & M & Admiration 72.7, Amusement 90.3, Anger 58.8, Annoyance 44.3, Approval 24.6, Caring 52.9, Confusion 73.2, Curiosity 64.2, Desire 35.7, Disappointment 58.0, Disapproval 52.3, Disgust 55.3, Embarrassment 30.4, Excitement 56.4, Fear 75.4, Gratitude 75.7, Grief 100.0, Joy 80.4, Love 86.8, Nervousness 83.3, Optimism 51.4, Pride 42.9, Realization 27.0, Relief 28.6, Remorse 38.6, Sadness 71.6, Surprise 62.1\\
 & L & Admiration 40.2, Amusement 84.9, Anger 52.7, Annoyance 40.2, Approval 36.0, Caring 36.5, Confusion 74.2, Curiosity 64.8, Desire 58.9, Disappointment 59.1, Disapproval 81.0, Disgust 38.2, Embarrassment 30.4, Excitement 61.8, Fear 73.8, Gratitude 88.4, Grief 100.0, Joy 84.8, Love 86.2, Nervousness 83.3, Optimism 65.4, Pride 71.4, Realization 41.6, Relief 42.9, Remorse 70.5, Sadness 67.6, Surprise 64.4\\
 \bottomrule
\end{tabular}
\caption{\label{tab:label-wise-lir} Label-wise percentage of times the test example's gold label is in $\ambiglst$ (as obtained from \base), where M and L refers to Flan-PaLM 2 sizes.}
\end{table*}

\subsection{Sorting Order with Predicted Probability Distribution Entropy}
Since we have the predicted probability distribution of ICL demonstrations, we tried to sort the ICL demonstrations by increasing entropy order. However, it doesn't consistently improve model performance, which is shown in \autoref{sort-entropy} and \ref{sort-entropy-diff}.

\begin{table}[!t]
\centering
\small
\begin{tabular}{lll|ll}\toprule
\multirow{2}{*}{} & \multicolumn{2}{l|}{EDOS} & \multicolumn{2}{l}{SST} \\
 & M & L & M & L \\\midrule
\textsc{Ambig}-4\\
\hspace{.5em}{\scriptsize \textsc{+gold}} & 50.2 & 62.9 & 51.6 & 55.8 \\
\hspace{.5em}{\scriptsize \textsc{+gold+mis}} & 51.2 & 62.7 & 53.0 & 57.0 \\
\hspace{.5em}{\scriptsize \textsc{+gold+mis+pred}} & 53.4 & 63.8 & 52.7 & 57.7 \\\midrule
\textsc{Ambig}-8\\
\hspace{.5em}{\scriptsize \textsc{+gold}} & 48.1 & 63.3 & 53.2 & 56.5 \\
\hspace{.5em}{\scriptsize \textsc{+gold+mis}} & 50.4 & 62.9 & 53.6 & 57.4 \\
\hspace{.5em}{\scriptsize \textsc{+gold+mis+pred}} & 50.3 & 62.9 & 54.3 & 57.1 \\\bottomrule
\end{tabular}
\caption{\label{sort-entropy} F1 macro scores (\%) of our method. M and L refers to size of Flan-PaLM 2. The ICL demonstrations are sorted by increased entropy order.}
\end{table}

\begin{table}[!t]
\centering
\small
\begin{tabular}{lll|ll}\toprule
\multirow{2}{*}{} & \multicolumn{2}{l|}{EDOS} & \multicolumn{2}{l}{SST} \\
 & M & L & M & L \\\midrule
\textsc{Ambig}-4\\
\hspace{.5em}{\scriptsize \textsc{+gold}} & ~0.9 & 0.3 & 0.1 & -0.3 \\
\hspace{.5em}{\scriptsize \textsc{+gold+mis}} & -1 & 1 & 0.7 & -0.4 \\
\hspace{.5em}{\scriptsize \textsc{+gold+mis+pred}} & -0.5 & 0.9 & -0.6 & -0.3 \\\midrule
\textsc{Ambig}-8\\
\hspace{.5em}{\scriptsize \textsc{+gold}} & 0.6 & 0.1 & 0.3 & 0 \\
\hspace{.5em}{\scriptsize \textsc{+gold+mis}} & 0 & 0.9 & 0.2 & -0.3 \\
\hspace{.5em}{\scriptsize \textsc{+gold+mis+pred}} & -0.6 & 0.2 & 0 & -0.1 \\\bottomrule
\end{tabular}
\caption{\label{sort-entropy-diff} The difference of F1 macro scores (\%) between the ``increased entropy order'' and the ``averaged over 3 random seeds''.}
\end{table}

\subsection{Example on SST for Comparison between \topretr and \textsc{Ambig-ICL}}

We list example demonstrations on SST where the model correctly predict the test example in our proposed method, but wrongly classify it in \topretr in \autoref{tab:icl_comparison_sst5}. 

\begin{table*}[!t]
    \centering
    \small
    \begin{tabular}{p{0.14\linewidth}p{0.65\linewidth}p{0.1\linewidth}}\toprule
    \multicolumn{3}{l}{\textbf{Test Example: A hip ride into hyper-time, Clockstoppers is a lively and enjoyable adventure for all ages at any time.}}\\
    \multicolumn{3}{l}{\textbf{$\mathbf{\ambiglst}$: Great, Good} \hfill \textbf{Gold label: Great}} \\\midrule
    \topretr & \begin{minipage}[t]{\linewidth}
    1. Bad: See \textit{Clockstoppers} if you have nothing better to do with 94 minutes.\\
    2. Bad: Time stands still in more ways that one in\textit{ Clockstoppers}, a sci-fi thriller as lazy as it is interminable. \\
    3. Bad: \textit{Clockstoppers} is one of those crazy, mixed-up films that doesn't know what it wants to be when it grows up.\\
    4. Good: Even with all its botches, Enigma offers all the pleasure of a handsome and well-made entertainment.
    \end{minipage} & Predicted: Good \\\midrule
    \multicolumn{3}{l}{\textsc{Ambig-ICL}}  \\
    \hspace{.5em}{\scriptsize\textsc{+gold}} & \begin{minipage}[t]{\linewidth}
    1. Good: Even with all its botches, Enigma offers all the pleasure of a handsome and well-made entertainment. \\
    2. Great: A breathtaking adventure \textit{for all ages}, Spirit tells its poignant and uplifting story in a stunning fusion of music and images.\\
    3. Great: A rollicking ride, with jaw-dropping action sequences, striking villains, a gorgeous color palette, astounding technology, stirring music and a boffo last hour that leads up to a strangely sinister happy ending.\\
    4. Great: This gorgeous epic is guaranteed to lift the spirits of the whole family.\\
    \end{minipage} & Predicted: Great\\
    \hspace{.5em}{\scriptsize \textsc{+gold+mis}}& \begin{minipage}[t]{\linewidth}
    1. Good: As action-adventure, this space-based homage to Robert Louis Stevenson's Treasure Island fires on all plasma conduits. \\
    2. Good: Horns and Halos benefits from serendipity but also reminds us of our own responsibility to question what is told as the truth.\\
    3. Great: Return to Never Land is reliable, standard Disney animated fare, with enough creative energy and wit to entertain \textit{all ages}.\\
    4. Great: It's a smart, solid, kinetically-charged spy flick worthy of a couple hours of summertime and a bucket of popcorn.\\
    \end{minipage} & Predicted: Great\\
    \hspace{.5em}{\scriptsize \textsc{+gold+mis+pred}} & \begin{minipage}[t]{\linewidth}
    1. Good: As action-adventure, this space-based homage to Robert Louis Stevenson's Treasure Island fires on all plasma conduits.\\
    2. Good: Horns and Halos benefits from serendipity but also reminds us of our own responsibility to question what is told as the truth.\\
    3. Great: Return to Never Land is reliable, standard Disney animated fare, with enough creative energy and wit to entertain all ages.\\
    4. Great: It's a smart, solid, kinetically-charged spy flick worthy of a couple hours of summertime and a bucket of popcorn.
    \end{minipage}  & Predicted: Great\\
    \bottomrule
    \end{tabular}
    \caption{\label{tab:icl_comparison_sst5} Example demonstrations selected by the \topretr and our proposed method \textsc{Ambig-ICL} for the SST task, for $n=4$. The model used here for prediction is Flan-PaLM 2 (L). 
    }
\end{table*}

\subsection{Responsible AI Checklist}

\paragraph{Packages for Evaluations.} We use SciPy \citep{2020SciPy-NMeth} and scikit-learn \citep{sklearn_api} for evaluations.

\end{document}